\documentclass[10pt,twocolumn,letterpaper]{article}

\usepackage{wacv}
\usepackage{times}
\usepackage{epsfig}
\usepackage{graphicx}
\usepackage{amsmath}
\usepackage{amssymb}
\usepackage{tabularx}
\usepackage{tikz}
\usepackage{booktabs} 
\usepackage{arydshln} 
\newcolumntype{R}[1]{>{\raggedleft\let\newline\\\arraybackslash\hspace{0pt}}m{#1}}

\def\wacvPaperID{635} 

\wacvfinalcopy 

\ifwacvfinal
\def\assignedStartPage{1} 
\fi

\usepackage{xr-hyper}

\makeatletter
\newcommand*{\addFileDependency}[1]{
  \typeout{(#1)}
  \@addtofilelist{#1}
  \IfFileExists{#1}{}{\typeout{No file #1.}}
}
\makeatother

\newcommand*{\myexternaldocument}[1]{%
    \externaldocument{#1}%
    \addFileDependency{#1.tex}%
    \addFileDependency{#1.aux}%
}

\myexternaldocument{suppl_material}
\ifwacvfinal
\usepackage[breaklinks=true,bookmarks=false]{hyperref}
\else
\usepackage[pagebackref=true,breaklinks=true,colorlinks,bookmarks=false]{hyperref}
\fi

\ifwacvfinal
\else
\fi

\begin{document}

\title{Stochastic-YOLO: Efficient Probabilistic Object Detection under Dataset Shifts}

\author{Tiago Azevedo\thanks{Work performed while the author was at Arm ML Research Lab}\\
Department of Computer Science and Technology\\
University of Cambridge\\
{\tt\small tiago.azevedo@cst.cam.ac.uk}
\and
Ren\'{e} de Jong\\
Arm ML Research Lab\\
{\tt\small rene.dejong@arm.com}
\and
Matthew Mattina\\
Arm ML Research Lab\\
{\tt\small matthew.mattina@arm.com}
\and
Partha Maji\\
Arm ML Research Lab\\
{\tt\small partha.maji@arm.com}
}

\maketitle

\begin{abstract}

In image classification tasks, the evaluation of models' robustness to increased dataset shifts with a probabilistic framework is very well studied. However, object detection (OD) tasks pose other challenges for uncertainty estimation and evaluation. For example, one needs to evaluate both the quality of the label uncertainty (i.e., \textit{what?}) and spatial uncertainty (i.e., \textit{where?}) for a given bounding box, but that evaluation cannot be performed with more traditional average precision metrics (e.g., mAP). In this paper, we adapt the well-established YOLOv3 architecture to generate uncertainty estimations by introducing stochasticity in the form of Monte Carlo Dropout (MC-Drop), and evaluate it across different levels of dataset shift. We call this novel architecture Stochastic-YOLO, and provide an efficient implementation to effectively reduce the burden of the MC-Drop sampling mechanism at inference time. Finally, we provide some sensitivity analyses, while arguing that Stochastic-YOLO is a sound approach that improves different components of uncertainty estimations, in particular spatial uncertainties.
   
\end{abstract}


\section{Introduction}

Current developments in computer vision are sustained by very powerful deep learning models, with proven capabilities to deliver good results. Due to its direct impact in daily human lives, it is of paramount importance to have models that generalise well to unseen cases during training time. In other words, these models need to be robust to dataset shifts and as a consequence there is the need to properly capture and evaluate their uncertainty measures.

This is well explored for image classification tasks, where it has been seen that many models get wrong very confidently as dataset shift increases~\cite{ovadia2019can}. But object detection (OD) tasks pose other types of challenges, as the metrics used in image classification tasks can be quite distinct. For example, in OD tasks, one has none to many bounding boxes being predicted for just a single image, in which \textit{uncertainty} semantics have a more diverse meaning. While in both image classification and OD tasks we can measure how certain the model is about \textit{what} the labels might be, in the OD case we also need to know how certain the model is about \textit{where} the objects (i.e., bounding boxes) are located. These two measures of uncertainty -- related to label and spatial information -- can be evaluated using a recently introduced metric named Probability-based Detection Quality (PDQ)~\cite{Hall2020} (see Section~\ref{subsec:pdq}).

We introduce Stochastic-YOLO, a novel architecture adapted from the well-established YOLOv3~\cite{redmon2018yolov3}, into which we add Monte Carlo Dropout (MC-Drop) sampling to introduce stochasticity in the predictions, which in turn generate uncertainty estimations. Although this paper uses YOLOv3 as a starting point, we argue that our framework can potentially be applied with minimum modifications in many other OD models to make them more robust to dataset shifts. Although there are alternatives to introduce stochasticity, MC-Drop is computationally light-weight and scales well at inference time. Ideally, a Bayesian neural network~\cite{Welling2011,blundell2015weight} would provide a fully probabilistic framework with more precise measures of uncertainty, but Bayesian nets have the downside of significant memory footprint. Likewise, an ensemble of models is known in the literature to produce better class labels and uncertainty predictions~\cite{ovadia2019can,Wei2018}, but it also brings an obvious memory footprint and expensive training times. We claim that using MC-Drop in OD tasks is the best trade-off between cost and robustness to dataset shifts under a probabilistic framework.

Similarly to previous works on image classification tasks, we will systematically evaluate how OD models generalise on increasing levels of corruptions, mainly in terms of label and spatial uncertainties. Although all these topics are somehow scattered throughout literature, to the best of our knowledge, no recent paper systematically evaluated label and spatial quality of OD techniques in increasing levels of dataset shifts in a single work, with inference time performance in mind. We also provide some sensitivity analysis on specific decisions for MC-Drop implementation which we did not find in the literature. We hope this paper can help researchers and practitioners better understand how these special cases influence performance and efficiency, such that one can easily extend our developments to any model deployed in real-world scenarios. Our contributions can be summarised in the following three main points:

\begin{enumerate}
    \item Introduction of an efficient caching mechanism for MC-Drop, effectively reducing sampling burden at inference time, which can be directly adapted to other models.
    \item Improvements on YOLOv3 architecture with an introduction of MC-Drop and sensitivity analysis over important hyperparameters;
    \item Leveraging probabilistic-based metrics (e.g., PDQ) to systematically evaluate model robustness for different levels of uncertainty (i.e. spatial and label qualities), while showing MC-Drop is a sound approach to improve PDQ;
\end{enumerate}

This work on probabilistic OD can be used directly in autonomous driving applications, but it is also useful for a multitude of other scenarios. Healthcare and, in specific, the medical imaging field, is a major example that can benefit from this type of work~\cite{kohl2018probabilistic}, in which uncertainty quantifications are still in their early ages~\cite{Lundervold2019}. The field of robotics is another obvious example, in which uncertainty measures can benefit robustness in unmanned aerial vehicles~\cite{wu2019delving}, as well as drone image segmentation~\cite{Ma2018} and drone-based object tracking~\cite{hamdi2020drotrack}.

\ifwacvfinal
The code used in this work is publicly available in the following Github link: \url{https://github.com/tjiagoM/stochastic-YOLO/}.
\else
The code used in this work is available in the following \textbf{anonymous} link: \url{https://github.com/anonymous-user-635/wacv_submission_635}.
\fi

\section{Related Work}
\label{sec:related}

The study and analysis of uncertainty measures has seen an increased interest thanks to Bayesian neural networks~\cite{kendall2017uncertainties}, while still being an active area of research with recent contributions for uncertainty quantification still being proposed~\cite{van2020uncertainty}. There are many attempts in the literature to bring stochasticity to OD models in order to capture their prediction confidence. Such attempts can be summarised in four categories: (1) directly learning to output Gaussian parameters for each bounding box coordinate, (2) using Bayesian approaches (e.g., Bayesian neural networks) to have a full probabilistic model, (3) using sampled-based Bayesian approximations like Monte Carlo Dropout (MC-Drop), and (4) using ensemble of models that generate a distribution of predictions, which can also be approximated as Gaussian parameters.

\paragraph{Uncertainty in Object Detection (OD):}
Gaussian YOLOv3~\cite{Choi_2019_ICCV} adapted the YOLOv3 architecture~\cite{redmon2018yolov3} and corresponding loss so the model could output Gaussian parameters instead of single, deterministic coordinates. This approach significantly reduced the false positive cases while keeping similar inference time when compared to YOLOv3. Similarly, He et al.~\cite{He2019} proposed a new loss named KL Loss to learn localisation uncertainty (i.e., variance) together with bounding boxes, which could then empower a voting scheme to select bounding boxes. Another work~\cite{Kraus2019} on a large scale automotive pedestrian dataset also reimplemented YOLOv3 in order to estimate epistemic and aleatoric uncertainty by using MC-Drop. 
Finally, another related work~\cite{Myojin2019} used the variance introduced by MC-Drop on a YOLOv3 architecture to measure spatial uncertainty; as a result, detections of lunar craters could be accepted or rejected on the basis of that variance. However, the spatial quality of those bounding boxes were not directly evaluated using specific quantitative metrics and these detections did not need to be performed in real-time nor in safety-critical situations like Stochastic-YOLO is targeted for.

\begin{figure*}[ht!]
\centering
\resizebox{\textwidth}{!}{%
\input{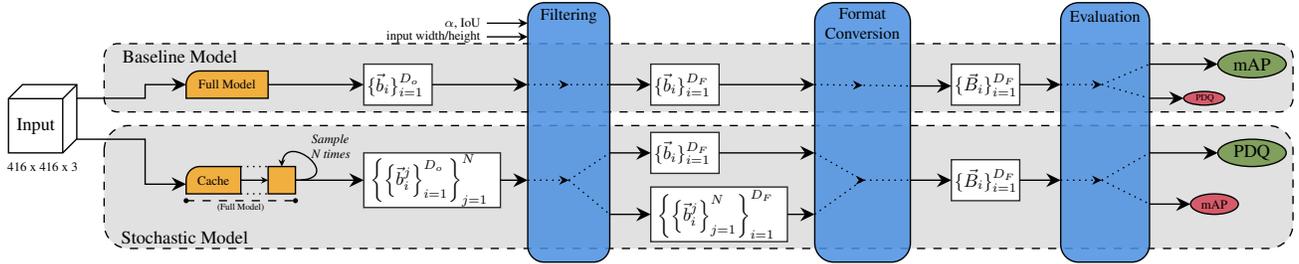}
}
\caption{Working blocks of a deterministic baseline model and a stochastic model with Monte Carlo Dropout, from which YOLOv3 and Stochastic-YOLO are special cases. A stochastic model outputs better probabilistic scores (i.e., PDQ) when compared to a deterministic baseline that outputs inflated mAP metrics. These inflated mAP metrics give a false sense of model performance. Illustration for confidence threshold $\alpha$, produced bounding box $\vec{b}$, number of bounding boxes originally produced $D_O$, final number of bounding boxes to be evaluated $D_F$, and converted bounding box $\vec{B}$. }
\label{fig:s_yolo_inference}
\end{figure*}

\paragraph{Robustness to non closed-set data:}
Improving the generalisability of neural networks has been a core component since early days, but only recently this issue has been more thoroughly evaluated in the case of dataset shifts or even completely out-of-distribution (OOD) datasets. Ovadia et al.~\cite{ovadia2019can} claim to have conducted the first rigorous, large-scale empirical comparison of predictive uncertainty for probabilistic deep learning methods under dataset shift. They evaluated both Bayesian and non-Bayesian methods, showing that both accuracy and the quality of the uncertainty consistently degrade as dataset shift is increased, regardless of the method. 
However, this study was conducted in an image classification task, leaving it open how these methods would behave in OD tasks. Indeed, some recent works analysed these issues in more detail for OD tasks. One work~\cite{Miller2018} claims to be the first to investigate the utility of dropout sampling for OD.
The authors were able to show that dropout variational inference is able to improve OD performance under open-set conditions when compared to non-bayesian models, even though they did not evaluated how good the spatial uncertainty was for their open-set datasets. Posterior works further extended this analysis: one work~\cite{Miller2018} compared different combinations of affinity measures and clustering methods for MC-Drop with a Single Shot Multi-Box Detector (SSD) in closed set data, near open-set data, and distant open-set data. 
Furthermore, some authors~\cite{Miller_2019_CVPR_Workshops} also analysed the performance of MC-Drop and Deep Ensembles in one- and two-stage OD models, showing Deep Ensembles yielding the best results. They have also introduced a new merging strategy that achieved comparative performance with only one hyperparameter needed, compared to three from other models, which we used in this paper (see Section~\ref{sec:methods}). Robustness to OOD cases can also be seen under the framework of adversarial attacks~\cite{zhang2019towards}; however, we willingly did not tackle that research direction in this work and focused instead on more systematic dataset shifts.

\paragraph{}
All these works showed advantages in considering localisation/spatial uncertainty in OD models for non closed-set data. Spatial uncertainty is many times used to help in deciding which bounding boxes are correct, but there was not an unified way to systematically evaluate measures of uncertainty for different open-set scenarios. Indeed, from the OD models presented in this section, only one~\cite{Miller_2019_CVPR_Workshops} used the unifying metric for probabilistic OD that we use in this paper (i.e., PDQ). Our work also has the objective of introducing stochasticity for good predictive uncertainty measures in a way that can be efficiently run at inference time, which we do not see in previous related literature.

\section{Methods}
\label{sec:methods}

\subsection{Stochastic-YOLO}

We introduce Stochastic-YOLO, a novel OD architecture based on YOLOv3~\cite{redmon2018yolov3} with efficiency in mind. We added dropout layers for Monte Carlo Dropout (MC-Drop) sampling in each one of the three YOLO heads; in specific, each head had a dropout layer before the last CNN layer, and before the second to last CNN layer. We argue that MC-Drop brings a good trade-off in terms of complexity and efficiency for introducing stochasticity at inference time. 

The introduction of dropout layers towards the end of a deep OD model, instead of after every layer, allows for an efficient sampling procedure. When sampling $N$ times from a deep OD model, one can cache the intermediate resulting feature tensor of one forward pass right until the first dropout layer. This cached tensor is deterministic (assuming that numerical errors are not significant), thus allowing only the last few layers of the model to be sampled from instead of making $N$ full forward passes through all the layers. Where a vanilla MC-Drop brings $O(N)$ time complexity for sampling $N$ times from a certain input, by caching the deterministic component this time complexity comes very close to what would be a deterministic model - $O(1)$ for small enough $N$ - corresponding to almost a single forward pass. We illustrate these gains at inference time for Stochastic-YOLO at Section~\ref{subsec:efficiency}.

For the specific case of Stochastic-YOLO, MC-Drop is applied at inference time by sampling 10 times from the model with the dropout layers activated. A common issue with one-stage detectors like YOLOv3 is that there will be a significant amount of overlapping bounding boxes being predicted; thus, it is necessary to apply non-maximum suppression (NMS) techniques to select only the best bounding boxes. The suppressions included in this work, both for Stochastic-YOLO and YOLOv3, consisted of (1) removing bounding boxes below a certain confidence threshold $\alpha$ (we tested for 0.1 and 0.5), (2) removing bounding boxes in which at least one coordinate was located outside image margins, and (3) removing bounding boxes with an intersection over union (IoU) above $0.6$ with another bounding box containing a higher classification score. These suppressions are calculated on the coordinates averaged from the 10 samples, thus closely following the Pre-NMS technique introduced by Miller et al.'s~\cite{Miller_2019_CVPR_Workshops}. There are other techniques in the literature for suppressing bounding boxes when using MC-Drop, but this one revealed to have a good trade-off between speed and performance when compared to other bounding box merging techniques~\cite{Miller_2019_CVPR_Workshops}. Indeed, the Pre-NMS technique makes the NMS inference time very similar between the non-stochastic and stochastic models, as the number of bounding boxes being compared is the same, with only the added time for averaging the sampled bounding boxes.

This entire process is summarised on a higher-level for non-stochastic and stochastic cases in Figure~\ref{fig:s_yolo_inference}, from which YOLOv3 and Stochastic-YOLO are special cases. For the non-stochastic baseline model at the top, its output is a set $ \{\vec{b}_{i} \}^{D_{o}}_{i=1}$, where $ \vec{b}_{i} \in \mathbb{R}^{5 + C} $ is a bounding box predicting $C$ possible labels, and $D_o$ is the number of bounding boxes originally produced by the specific model (e.g., YOLOv3). Each bounding box contains $5 + C$ real values: four representing the bounding box (i.e., x and y coordinates, and width and height), one representing the objectness score (i.e., softmax score that the bounding box contains an object), and $C$ softmaxed scores for each possible label. This set enters the \textit{Filtering} block which contains the suppression techniques described in the previous paragraph, now producing a smaller set  $ \{\vec{b}_{i} \}^{D_{F}}_{i=1}$ of bounding boxes, where $D_F$ is the final number of bounding boxes to be evaluated. For the stochastic model, the number of bounding boxes being originally produced is instead $N \times D_o$, where $N$ is the number of MC-Drop samples. This distinction in the stochastic model's output, when compared to the non-stochastic one, will make the \textit{Filtering} block to have an extra output: for each averaged bounding box $\vec{b}_{i}$ not being filtered, we need the corresponding $N$ samples of that bounding box, represented as $\left\{\vec{b}^{j}_{i}\right\}^{N}_{j=1}$.

A further \textit{Format Conversion} block is needed to transform these sets into a format which can be evaluated from a probabilistic perspective in the \textit{Evaluation} block (see Section~\ref{subsec:pdq}). In practice, a bounding box vector

\begin{align}
    \vec{b}_i = \left [ x, y, width, height, \mathrm{obj}, p_0, ..., p_C \right ]
\end{align}

is transformed into

\begin{align}
    \vec{B}_i = \left [ \bar{x}_1, \bar{y}_1, \bar{x}_2, \bar{y}_2, \Sigma^1, \Sigma^2, \hat{p}_0, ..., \hat{p}_C \right ],
\end{align}

where now we have two coordinates for the top-left corner (i.e., $(\bar{x}_1, \bar{y}_1)$) and bottom-right corner (i.e., $(\bar{x}_2, \bar{y}_2)$) instead of a single coordinate with width/height. In this work, for a deterministic model these values correspond to those originally in $\vec{b}_i$, whereas for the stochastic model these are the average coordinates across the $N$ samples. Each softmaxed score is transformed such that $\hat{p}_i = p_i \times \mathrm{obj}$. Finally, $\Sigma^i$ is a covariance matrix for one coordinate with the following representation:

\begin{align}
    \Sigma^i =
        \begin{bmatrix}
        \Sigma^i_{xx} & \Sigma^i_{xy}\\ 
        \Sigma^i_{yx} & \Sigma^i_{yy}
        \end{bmatrix},
    \label{eq:covars_matrix}
\end{align}

in which for the deterministic model these will all be zeros, whereas for the stochastic model each covariance matrix is calculated from the distribution of the $N$ sampled points in each coordinate. When a covariance matrix is not positive semi-definite, we transform it by computing the eigen decomposition, and reconstructing the matrix again with eigenvalues of zero where they were previously negative.

We highlight that the reasoning in this section illustrated in Figure~\ref{fig:s_yolo_inference} can be applied in almost any OD model that outputs some set of bounding boxes, even though in our experiments we illustrate this for YOLOv3 and Stochastic-YOLO only. Indeed, researchers might want to consider other merging strategies for their specific needs~\cite{Miller2019}.

\subsection{Systematic Evaluation of Dataset Shift Scenarios}

We used the Python package proposed by Michaelis et al.~\cite{michaelis2019dragon} to systematically evaluate the robustness of our models to increasing dataset shifts. Among other contributions, the package provides a total of 5 severity levels across 15 corruption types to be applied in images, some of them corresponding to distortions that can be seen more often in real-world scenarios, like rain, snow, and fog. From these 15 corruptions, we used all but glass blur due to the extended time taken to be calculated, when compared with all the others.

Michaelis et al.~\cite{michaelis2019dragon} also presented a single measure to evaluate relative performance degradation under corruption, named relative performance under corruption (rPC):

\begin{equation}
\textrm{rPC} = \frac{\frac{1}{N_c} \sum\limits_{c=1}^{N_c} \frac{1}{N_s} \sum\limits_{s=1}^{N_s} P_{c,s}}{P_{\textrm{clean}}},
\end{equation}

where $P_{c,s}$ is the performance of the model for severity level $s$ on corruption $c$, $N_c$ is the number of corruptions applied (i.e., 14 in our case) and $N_s$ the number of severity levels (i.e., 5 in our case). $P_{\textrm{clean}}$ corresponds to the performance of the model without any corruption in the dataset. Note that in the original paper the authors only used mAP as a performance metric, but this can be applied in any performance metric. For instance, $\textrm{rPC}_{m_i}$ corresponds to the relative performance of the model under corruption for metric $m_i$.

\subsection{Probability-based Detection Quality (PDQ)}
\label{subsec:pdq}

OD models have been predominantly evaluated using some variant of average precision (AP) in research papers and competitions likewise. This type of metric is not able to systematically evaluate uncertainty measures that are getting prevalent in OD tasks (see Section~\ref{sec:related}). We use the Probability-based Detection Quality (PDQ) metric~\cite{Hall2020}, which is calculated between the set of ground truths $G$, and the set of detections $D$. This metric is built on top of two ideas: label quality and spatial quality.

Label quality $Q_L$ on the $f$-th image (named \textit{frame} instead of \textit{image} in the original paper) between a $i$-th ground-truth object $G^f_i$ and a $j$-th detection $D^f_j$, can be defined as:

\begin{equation}
Q_L\left ( G^f_i, D^f_j \right ) = \mathbf{I}_j^f\left ( \hat{c}_i^f \right ),
\end{equation}

where $\mathbf{I}_j^f$ is the probability distribution across all classes on the $j$-th detection object, and $\hat{c}_i^f$ the ground-truth class. Note that this ignores whether the class is the highest ranked in the probability distribution $\mathbf{I}_j^f$, thus effectively evaluating the quality of \textit{what} object is being predicted.

Spatial quality $Q_S$ on the $f$-th image between a $i$-th ground-truth object $G^f_i$ and a $j$-th detection $D^f_j$, can be defined as:

\begin{align}
    &Q_S\left ( G^f_i, D^f_j \right ) = \nonumber \\
    &\mathrm{exp} \left ( - \left ( L_{FG}\left (G^f_i, D^f_j \right ) + L_{BG}\left (G^f_i, D^f_j \right ) \right ) \right ),
\end{align}

where $L_{FG}$ and $L_{BG}$ are two loss terms for the foreground and background of that image, respectively. This spatial quality will be equal to 1 (i.e., maximum value) when the detector $D^f_j$ assigns a probability of 1 to all the ground-truth pixels in $G^f_i$, while not assigning any probability to the other pixels (i.e., background). The probability distribution used to measure these two losses is calculated by joining the two covariance matrices from each corner introduced in Equation~\ref{eq:covars_matrix}. For more details on how this is calculated we refer the reader to the original paper~\cite{Hall2020}.

By calculating the geometric mean of these two terms we have the pairwise PDQ (pPDQ):

\begin{equation}
    \mathrm{pPDQ}(G^f_i, D^f_j) = \sqrt{Q_S\left ( G^f_i, D^f_j \right ) \cdot Q_L\left ( G^f_i, D^f_j \right )}.
\end{equation}

Note that the geometric mean will give higher values only when both terms are higher. Finally, by storing the pPDQs of all the non-zero assignments of the $f$-th image in a vector $\mathbf{q}^f=\left [ q_1^f, ..., q_{N^f_{TP}}^f \right ]$, with $N^f_{TP}$ being the number of true positives, we can define the final PDQ score as a summation over all those pPDQ scores, averaged across all true positives, false negatives, and false positives:

\begin{align}
    &PDQ(G,D) = \nonumber \\
    &\frac{1}{\sum\limits_{f=1}^{N_f}\left ( N_{TP}^f + N_{FN}^f + N_{FP}^f \right )} \sum\limits_{f=1}^{N_F} \sum\limits_{i=1}^{N_{TP}^f}\mathbf{q}^f(i),
\end{align}

where $N_F$ is the number of images being evaluated. This metric can effectively evaluate in a single value both label and spatial uncertainties of an OD model. Furthermore, another advantage of this metric is the flexibility that can bring to evaluations: from a single PDQ value, to more granular insights about spatial and label uncertainties, which we analyse in this paper.

\section{Experiments and Results}

\subsection{Training Procedure}
\label{subsec:training}
We have used and adapted Ultralytics' open-source implementation\footnote{\url{https://github.com/ultralytics/yolov3}} of YOLOv3 for Pytorch~\cite{pytorch2019} with input size of $416 \times 416$. Training followed all the repository's default hyperparameters, unless stated otherwise. Baseline YOLOv3 model was trained for 200 epochs using Stochastic Gradient Descent, cosine learning rate decay~\cite{loshchilov2016}, and data augmentation. Final model is selected according to best mAP in the validation set. Training and evaluation was performed using the \textit{train2017} and \textit{val2017} splits provided by MS COCO~\cite{Lin2014} 2017 release, with 118,287 and 5,000 images, respectively. Stochastic-YOLO is created by inserting the dropout layers in the best YOLOv3 model as previously described. For comparison, we train an ensemble of five YOLOv3 models, where each one was trained in the same way with different random seeds when initialising the network's weights.

YOLOv3 baseline model was trained with a batch size of 20, distributed across 2 \textit{GeForce RTX 2080} GPUs, producing training times of around 48 minutes per epoch. Ensemble models were trained with a batch size of 40, distributed across 2 \textit{GeForce RTX 2080 Ti} GPUs, producing training times of around 35 minutes per epoch. When Stochastic-YOLO is fine-tuned, this is done under the same conditions as YOLOv3, but with dropout layers activated at training time and for 50 extra epochs.

\subsection{Robustness to Corruptions}

Table~\ref{tab:model_results} summarises metrics and their relative performance under corruption (rPC) across different models. For every model we show two confidence thresholds -- 0.1 and 0.5 -- and for all the models a confidence threshold of 0.1 corresponds to higher values of mAP and better robustness for mAP. However, for the label and spatial quality metrics (and corresponding relative performances under corruption), all the models perform significantly worse when a confidence threshold of 0.1 is used. This illustrates well-known concerns in the field, where many developers try to decrease the confidence threshold to inflate mAP values, giving a false sense of good performance. For more detailed results, see Supplemental Table~S1.

\newcommand{\ra}[1]{\renewcommand{\arraystretch}{#1}}
\setlength\dashlinedash{0.5pt}
\setlength\dashlinegap{1.5pt}

\begin{table*}[h!]
\ra{1.1}
\centering
\caption{Overall results across different models and metrics, where Lbl and Sp mean average label and spatial uncertainty quality, respectively. In parenthesis confidence threshold. S-YOLO means Stochastic-YOLO in which the corresponding number is the dropout percentage applied, and -X means fine-tuned model. In bold best results for each metric.}
\label{tab:model_results}
\resizebox{\textwidth}{!}{%
\begin{tabular}{@{}lR{1.3cm}R{1.7cm}lrrlrrlrr@{}}
\toprule
 & \multicolumn{2}{c}{\textbf{Deterministic Metrics (\%)}} && \multicolumn{8}{c}{\textbf{Probabilistic Metrics (\%)}}\\ \cmidrule{2-3} \cmidrule{5-12}
 
  & \multicolumn{1}{r}{mAP} & \multicolumn{1}{c}{$\textrm{rPC}_{mAP}$} & \phantom{a} & \multicolumn{1}{c}{PDQ} & \multicolumn{1}{c}{$\textrm{rPC}_{PDQ}$} & \phantom{a} & \multicolumn{1}{c}{Lbl} & \multicolumn{1}{c}{$\textrm{rPC}_{Lbl}$} & \phantom{a} & \multicolumn{1}{c}{Sp} & $\textrm{rPC}_{Sp}$\\
 
 \toprule
YOLOv3 (0.1) & 34.43 & 20.64 && 7.19 & 4.88 && 55.27 & 48.38 && 14.53 & 12.18 \\
YOLOv3 (0.5) & 26.07 & 14.64 && 9.26 & 5.40 && \textbf{72.88} & \textbf{69.70} && 18.42 & 16.11 \\ \hdashline
Ensemble-5 (0.1) & \textbf{37.14} & \textbf{22.28} && 18.36 & \textbf{13.38} && 50.66 & 43.25 && 39.24 & 37.66 \\
Ensemble-5 (0.5) & 26.00 & 14.07 && 19.53 & 11.50 && 72.21 & \textbf{69.63} && 49.21 & 48.59 \\ \hdashline
S-YOLO-25 (0.1) & 31.67 & 19.31 && 17.73 & \textbf{13.40} && 48.18 & 41.67 && 37.27 & 34.84 \\
S-YOLO-25 (0.5) & 22.72 & 12.91 && \textbf{20.27} & 12.46 && 70.17 & 67.70 && 45.78 & 44.95 \\ \hdashline
S-YOLO-25-X (0.1) & 33.16 & 19.66 && 17.29 & 11.95 && 51.01 & 44.62 && 38.59 & 34.58 \\
S-YOLO-25-X (0.5) & 24.20 & 13.58 && 19.08 & 11.39 && 70.99 & 68.09 && 48.76 & 45.58 \\ \hdashline
S-YOLO-75 (0.1) & 17.36 & 11.11 && 4.44 & 4.20 && 29.88 & 24.79 && 19.71 & 17.79 \\
S-YOLO-75 (0.5) & 7.47 & 4.36 && 7.81 & 5.09 && 61.32 & 60.71 && 25.88 & 26.61 \\ \hdashline
S-YOLO-75-X (0.1) & 29.36 & 17.63 && 17.86 & 12.57 && 42.94 & 38.68 && 39.36 & 36.82 \\
S-YOLO-75-X (0.5) & 17.76 & 10.11 && 15.20 & 9.51 && 66.34 & 64.41 && \textbf{52.27} & \textbf{49.97} \\
\bottomrule
\end{tabular}
}
\vspace{-4mm}
\end{table*}

\raggedbottom

Fine-tuning Stochastic-YOLO models usually brings better metrics and robustness to corruptions when compared to Stochastic-YOLO used directly from a pre-trained YOLOv3 model with inserted dropout layers and no fine-tuning. Although for a dropout rate of 25\% these are just slight improvements, for a dropout rate of 75\% they are much more substantial, even resulting in the best spatial quality and robustness for spatial quality among all models.

PDQ score, spatial quality and robustness for these two metrics more than doubled for Stochastic-YOLO model with 25\% dropout rate when compared to YOLOv3. At the same time, label quality, robustness for label quality, and robustness for mAP only reduced around 2\% when compared to YOLOv3 with the same 0.5 confidence threshold. The mAP metric is the only one that is more significantly impacted (almost 4\%) with Stochastic-YOLO for a confidence threshold of 0.5 at 25\% dropout rate, but the issues with mAP usefulness are well-known in the field.

It is noteworthy the very good label quality across all models when a confidence threshold of 0.5 is used. The disproportionate capability of YOLOv3 to produce very high label quality measures but very low spatial quality measures has been previously seen in the literature~\cite{Hall2020}. This is also the performance metric more robust to corruptions, with a relative performance (i.e., $\textrm{rPC}_{Lbl}$) of almost 70\% in the best models. By using a confidence threshold of 0.1 instead of 0.5, these two metrics are the most severely impacted.

\subsection{Model Fine-tuning}

\begin{figure}[h]
\centering
\includegraphics[width=.95\linewidth]{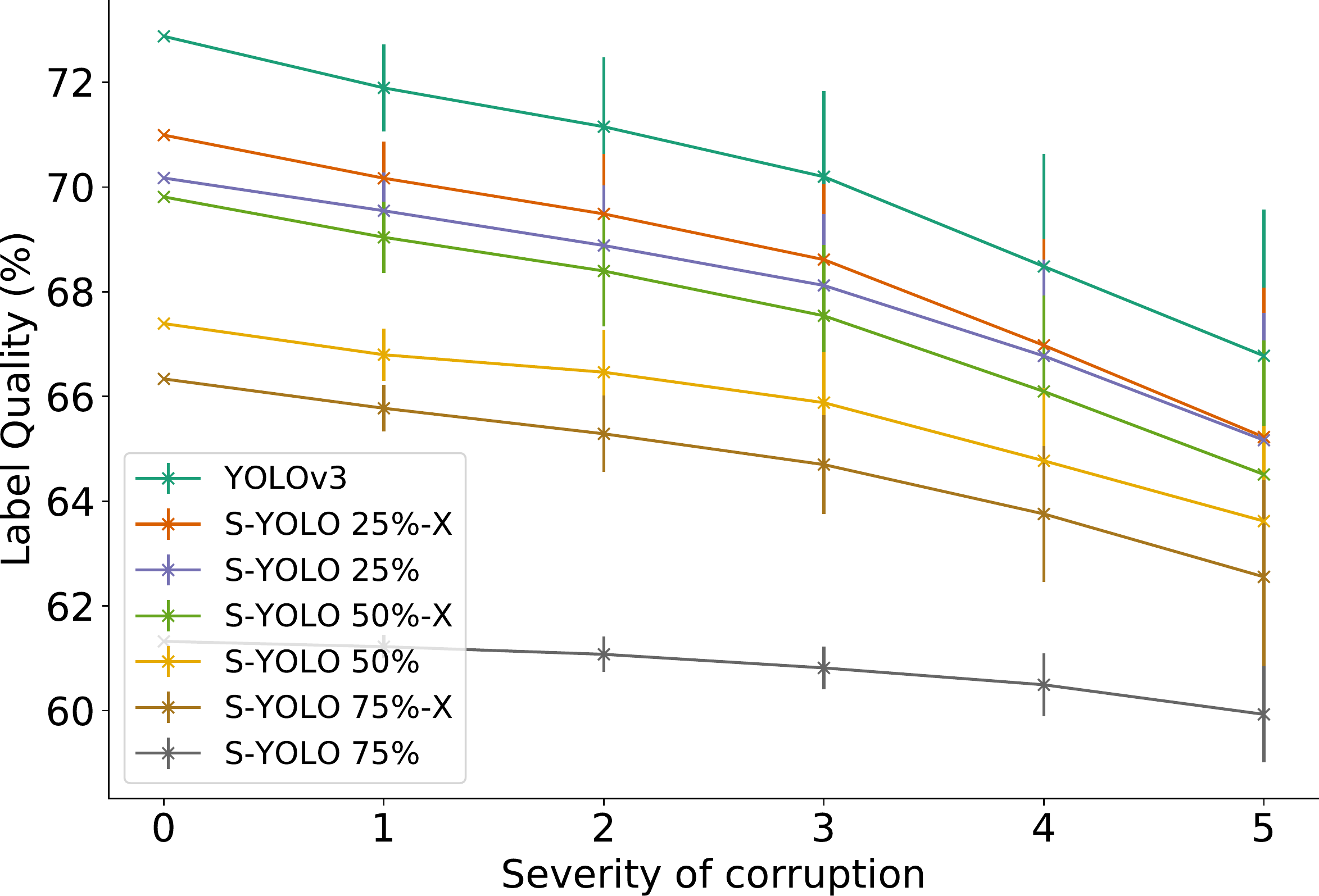}
\caption{Label Quality for different models, averaged across all corruptions for each severity, with 0.5 confidence threshold. Error bars correspond to one standard deviation. S-YOLO means Stochastic-YOLO followed by dropout rate. -X means fine-tuned model. Note the reduced variation in the y-axis.}
\label{fig:finetune_avg_label}
\end{figure}

\begin{figure}[h]
\centering
\includegraphics[width=.95\linewidth]{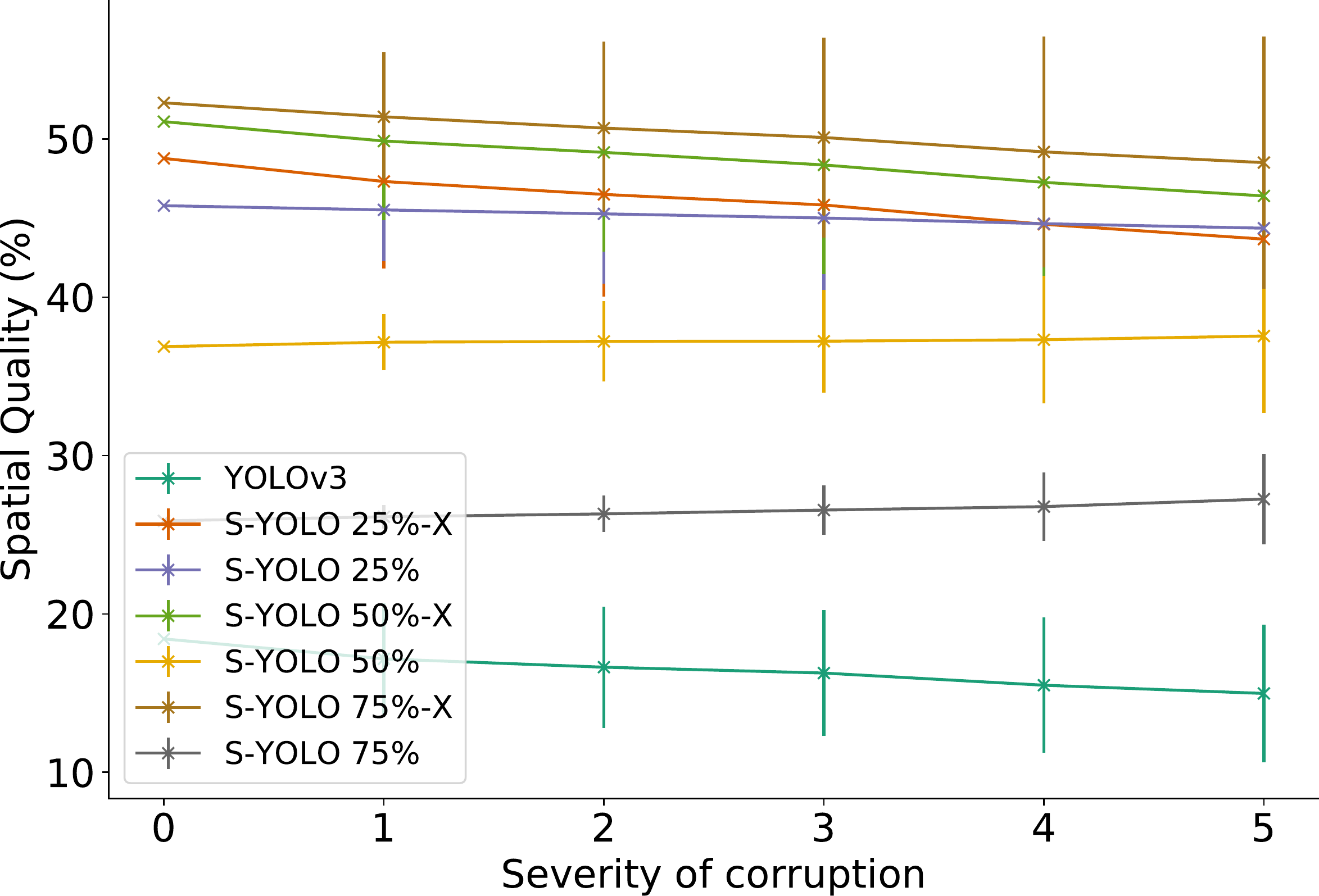}
\caption{Spatial Quality for different models, averaged across all corruptions for each severity, with 0.5 confidence threshold. Error bars correspond to one standard deviation. S-YOLO means Stochastic-YOLO followed by dropout rate. -X means fine-tuned model.}
\label{fig:finetune_avg_spatial}
\end{figure}

As it can be seen in Table~\ref{tab:model_results}, the decision of further fine-tuning Stochastic-YOLO can impact overall results. We check three representative dropout rates (25\%, 50\%, and 75\%) and resulting spatial and label qualities over corruptions averaged across dataset shifts. These results are depicted in Figure~\ref{fig:finetune_avg_label} and Figure~\ref{fig:finetune_avg_spatial} for label and spatial quality, with a confidence threshold of 0.5. The same plots can be seen for the PDQ and mAP scores in supplementary material (Figures~S1 and~S2).

Both uncertainty qualities (i.e., label and spatial) follow similar patterns: fine-tuning Stochastic-YOLO produces better averaged metrics across increased levels of dataset shifts, with these improvements being smaller for a smaller dropout rate (i.e., 25\%), and more noticeable for a higher dropout rate (i.e., 75\%). For spatial quality this behaviour is even more evident, as the best averaged spatial qualities are produced when fine-tuning with 75\% dropout rate, followed by 50\% and 25\%. All these three fine-tuned models are the best, on average, for most cases when compared to all the other non fine-tuned models.

Nevertheless, Stochastic-YOLO with a dropout rate of 25\% with no fine-tuning seems to yield the best trade-off in terms of performance and complexity, as it achieves comparable results without the need to spend further time fine-tuning it.

\subsection{Sensitive Analysis on Dropout Rate}

\begin{figure}[h]
\centering
\includegraphics[width=.95\linewidth]{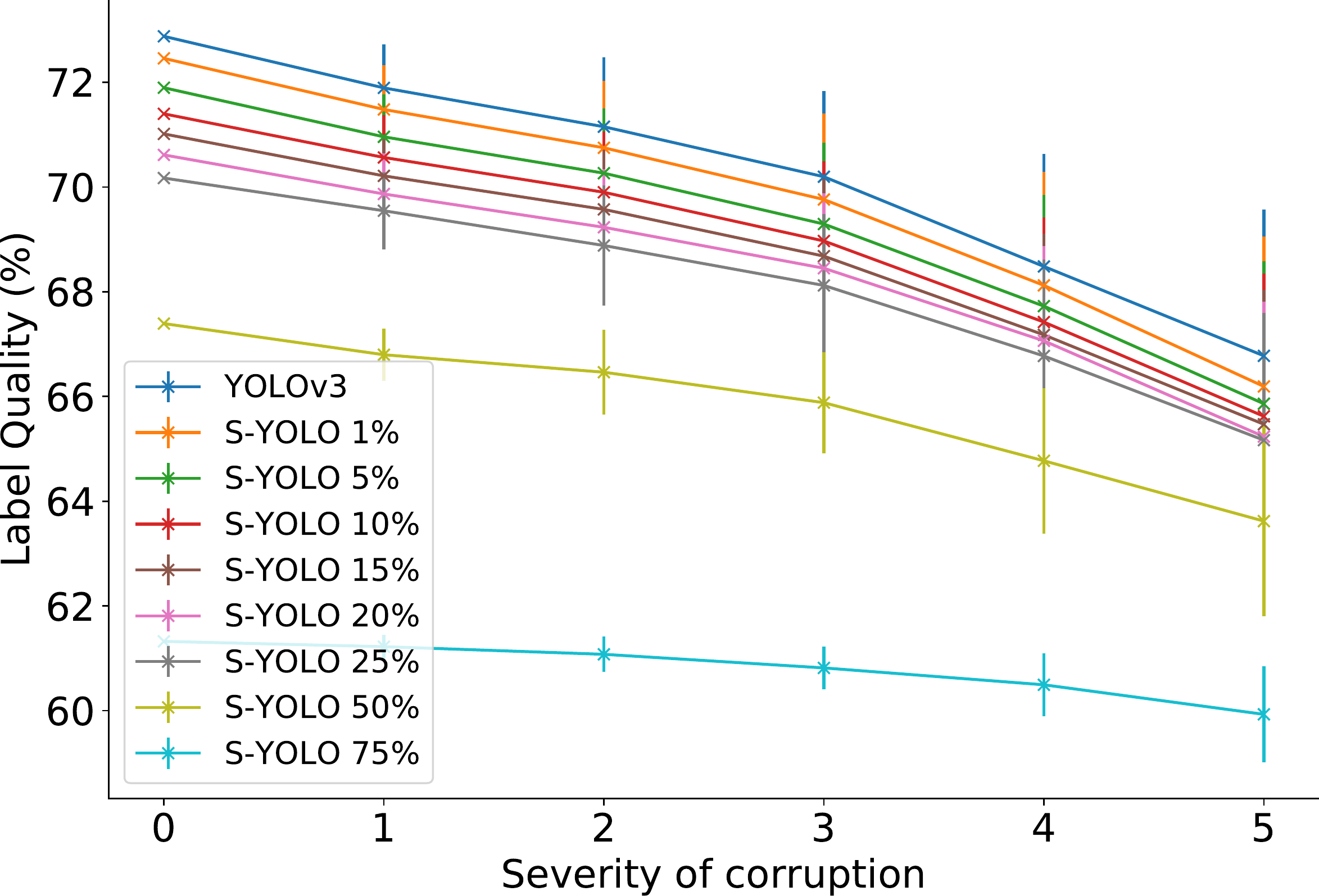}
\caption{Label Quality across different dropout rates, averaged across all corruptions for each severity, with 0.5 confidence threshold. Error bars correspond to one standard deviation. S-YOLO means Stochastic-YOLO. Note the reduced variation in the y-axis.}
\label{fig:droprates_avg_label}
\end{figure}

\begin{figure}[h]
\centering
\includegraphics[width=.95\linewidth]{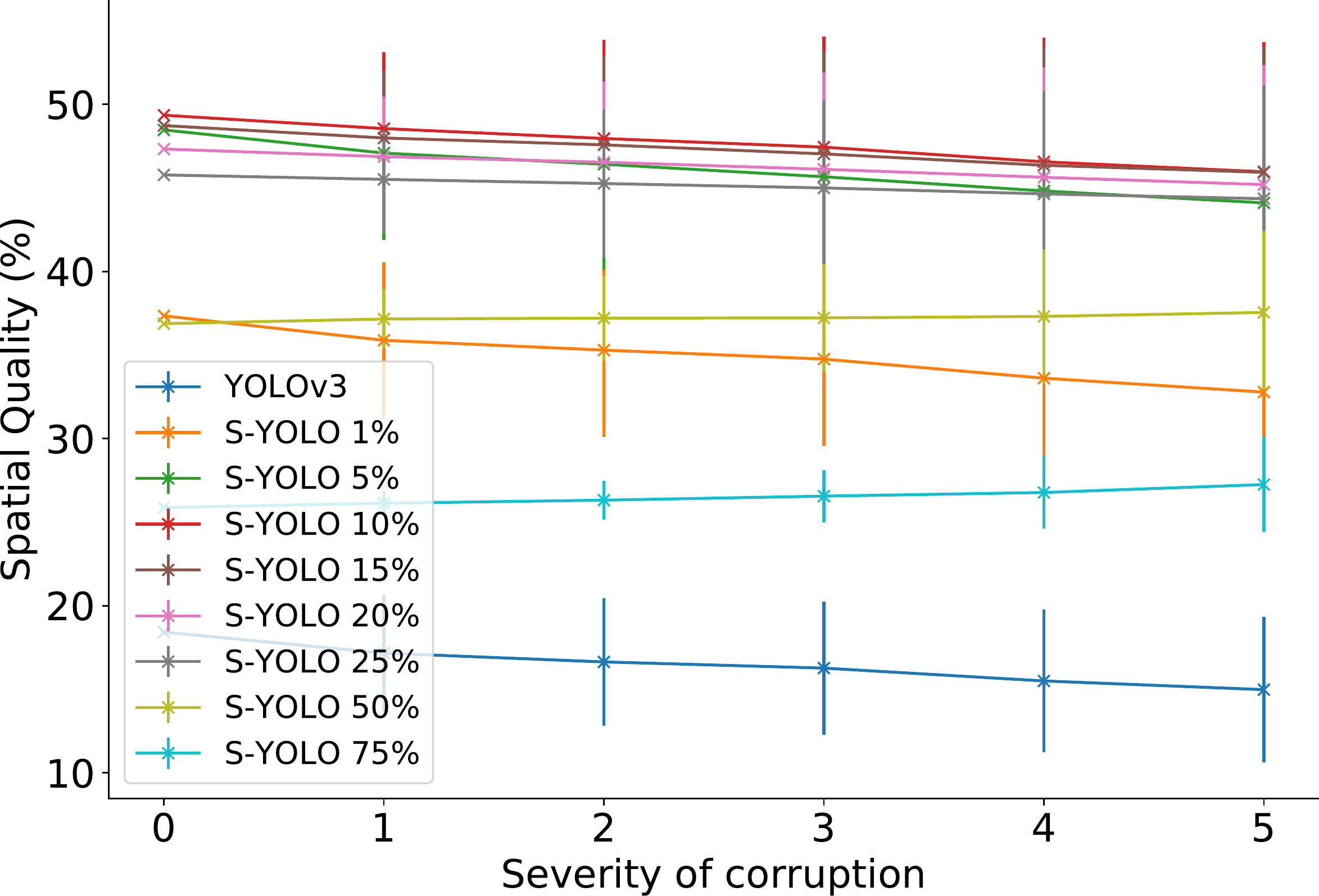}
\caption{Spatial Quality across different dropout rates, averaged across all corruptions for each severity, with 0.5 confidence threshold. Error bars correspond to one standard deviation. S-YOLO means Stochastic-YOLO.}
\label{fig:droprates_avg_spatial}
\end{figure}

We provide a sensitivity analysis on the influence of dropout rate, a hyperparameter that needs to be defined when using Stochastic-YOLO. Due to resource and time constrains, we focus this sensitivity analysis on Stochastic-YOLO without any further fine-tuning. Figure~\ref{fig:droprates_avg_label} and Figure~\ref{fig:droprates_avg_spatial} contain the results for label and spatial quality, with a confidence threshold of 0.5. As previously, the same plots can be seen for the PDQ and mAP scores in supplementary material (Figures~S3 and~S4).

The behaviour of Stochastic-YOLO is very consistent for label quality across increased dataset shifts: the smallest dropout rate produces the best label quality (only surpassed by YOLOv3), and consistently worsens for higher dropout rates. However, for the spatial quality the behaviour is not the same. There is a range of very similar results between 5\% and 25\%, with significant worse results below and above this range. Indeed, on average, with 1\% dropout rate Stochastic-YOLO yields worse results when compared to 50\% for spatial quality. Once again, Stochastic-YOLO with 25\% dropout achieves a good trade-off across different metrics without fine-tuning.

\subsection{Model Efficiency}
\label{subsec:efficiency}

To establish that the proposed improvement is indeed a more efficient solution, we need to validate Stochastic-YOLO's behaviour using real hardware. We measure the impact of the networks on inference time and power usage, by setting up an experiment running the different models on two different real devices. For this, we make use of the NVIDIA Jetson TX2 (TX2) platform and the NVIDIA GeForce RTX 2080 (RTX 2080) graphics card. Both platforms allow for an accurate power measurement during runtime. The TX2 has a Dual core Denver 2 CPU (2GHz) and a Quad-Core Arm Cortex-A57 CPU (2GHz). It comes with a 256-core NVIDIA Pascal GPU, and 8GB LPDDR4 memory. The RTX 2080 is a Turing architecture GPU with 2944 CUDA cores and 8GB GDDR6 memory. Each network inference is run 1000 times and throughout the run we measure the power. At the end of each experiment we calculate the average power consumption and inference times of each network.

\begin{figure}[h]
\includegraphics[width=.95\linewidth]{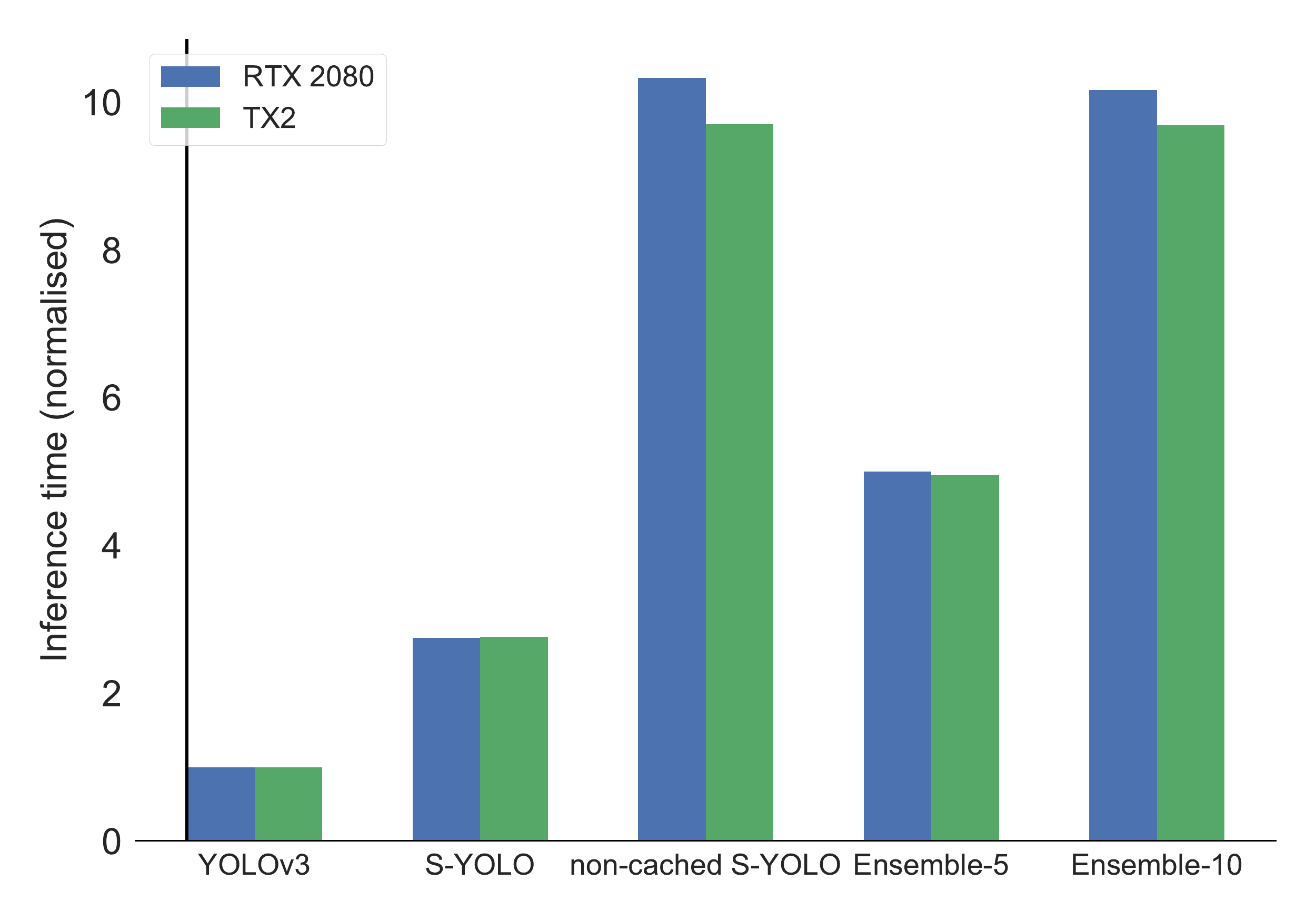}
\caption{Comparison of inference time of the different models normalised to the inference time of YOLOv3 on that platform, as run on a NVIDIA RTX 2080, and a NVIDIA Jetson TX2.}
\label{fig:performance_bar}
\end{figure}

\begin{figure}[h]
\includegraphics[width=.95\linewidth]{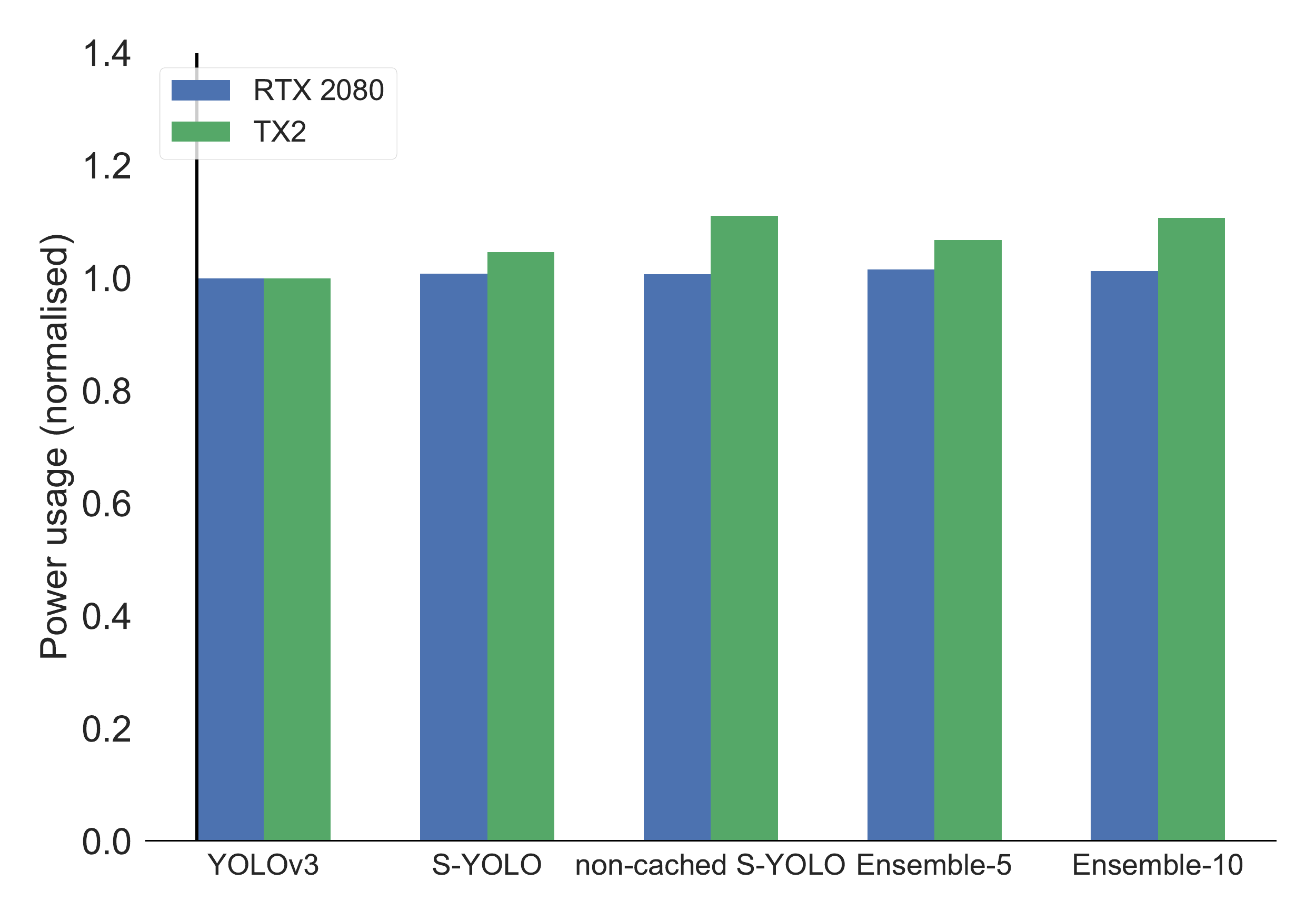}
\caption{Comparison of power usage of the different models normalised to the power usage of YOLOv3 on that platform; as run on a NVIDIA RTX 2080, and a NVIDIA Jetson TX2.}
\label{fig:power_bar}
\end{figure}

By performing measurements on both platforms, we illustrate the effect of the different model performances for high-performance systems as well as energy-efficient systems. The results of these measurements are summarised in Figure~\ref{fig:performance_bar} and Figure~\ref{fig:power_bar}. All the results are normalised to the results as measured for the YOLOv3 run on that platform. The measured inference time for YOLOv3 is 12~ms per instance on the RTX 2080, and 162~ms on the TX2. Its power usage is 216~W and 9~W for the RTX 2080 and the TX2, respectively. ``Ensemble-10'' was created only for comparison purposes, and corresponds in practice to duplicating the models from ``Ensemble-5''.

We see that the baseline run of YOLOv3 shows in all cases the lowest inference times, and the non-cached Stochastic-YOLO and the Ensemble show the highest. Since this non-cached version of Stochastic-YOLO is run ten times, we see an unsurprising ten-time slowdown in the inference times. However, the overhead for the cached model only shows a two to three times slowdown compared to the baseline and a three to four times speedup compared with the non-cached Stochastic-YOLO. The power consumption does increase only marginally in comparison with YOLOv3. This demonstrates that Stochastic-YOLO's caching mechanism is an efficient alternative to the full Stochastic-YOLO without any caching mechanism.

\section{Discussion}

In this work, we enhanced the well-established YOLOv3 architecture in order to generate uncertainty estimations in OD tasks. These improvements included the addition of Monte Carlo Dropout (MC-Drop), thus successfully generating both label and spatial uncertainties. The implementation of this architecture was achieved using a caching mechanism that enabled minimal impact on inference time when deploying the model in the real world. All these developments allowed for better resulting probabilistic metrics (i.e., PDQ), which have proven advantages over the more traditional mAP metric that can sometimes give a false sense of performance.

Although in the literature the use of MC-Drop usually includes dropout layers activated at both training and test times, in many OD architectures dropout is not a typical choice of regulariser for training. As a consequence, we developed Stochastic-YOLO in a way that can fit most of OD researchers current pipelines, in the sense that we show that direct application of MC-Drop in pre-trained models without any fine-tuning results in significant improvements (see Table~\ref{tab:model_results}). To further help the community decide how to use Stochastic-YOLO, we provided sensitivity analysis on dropout rate and on the decision of further fine-tuning the model, which to the best of our knowledge has not yet been fully explored in the field. 
Overall, we found that not fine-tuning Stochastic-YOLO yields sufficiently good results already, in which case a smaller dropout rate should be picked. For those with the time and computational resources for further fine-tuning their stochastic models with MC-Drop, a higher dropout rate should be picked instead. We hope that this work encourages other researchers to extend the ideas presented in this paper on their own models (as illustrated in Figure~\ref{fig:s_yolo_inference}) with the help of our publicly available code.

We identify four promising directions for future work that we are currently exploring: (1) conduct sensitivity analysis on the number and positions of dropout layers, as well as number of samples needed, such that it does not negatively impact efficiency, (2) further study more complex mechanisms in the hardware-level for efficient sampling, like scheduling techniques, (3) study the impact of other merging techniques in a way that does not negatively impair inference time, and (4) analyse how these systematic evaluations on probabilistic metrics and runtime will behave when stochasticity is brought to other models (e.g., Fast R-CNN and SSD).

{\small
\bibliographystyle{ieee_fullname}
\bibliography{wacv_bib}
}

\end{document}


\title{SUPPLEMENTARY MATERIAL\\Stochastic-YOLO: Efficient Probabilistic Object Detection under Dataset Shifts\vspace{-1.5em}}

\author{Tiago Azevedo\thanks{Work performed while the author was at Arm, Ltd.}\\
Department of Computer Science and Technology\\
University of Cambridge\\
{\tt\small tiago.azevedo@cst.cam.ac.uk}
\and
Ren\'{e} de Jong\\
Arm ML Research Lab\\
{\tt\small rene.dejong@arm.com}
\and
Matthew Mattina\\
Arm ML Research Lab\\
{\tt\small matthew.mattina@arm.com}
\and
Partha Maji\\
Arm ML Research Lab\\
{\tt\small partha.maji@arm.com}
}

\maketitle

   



\begin{figure}[h]
\centering
\includegraphics[width=.95\linewidth]{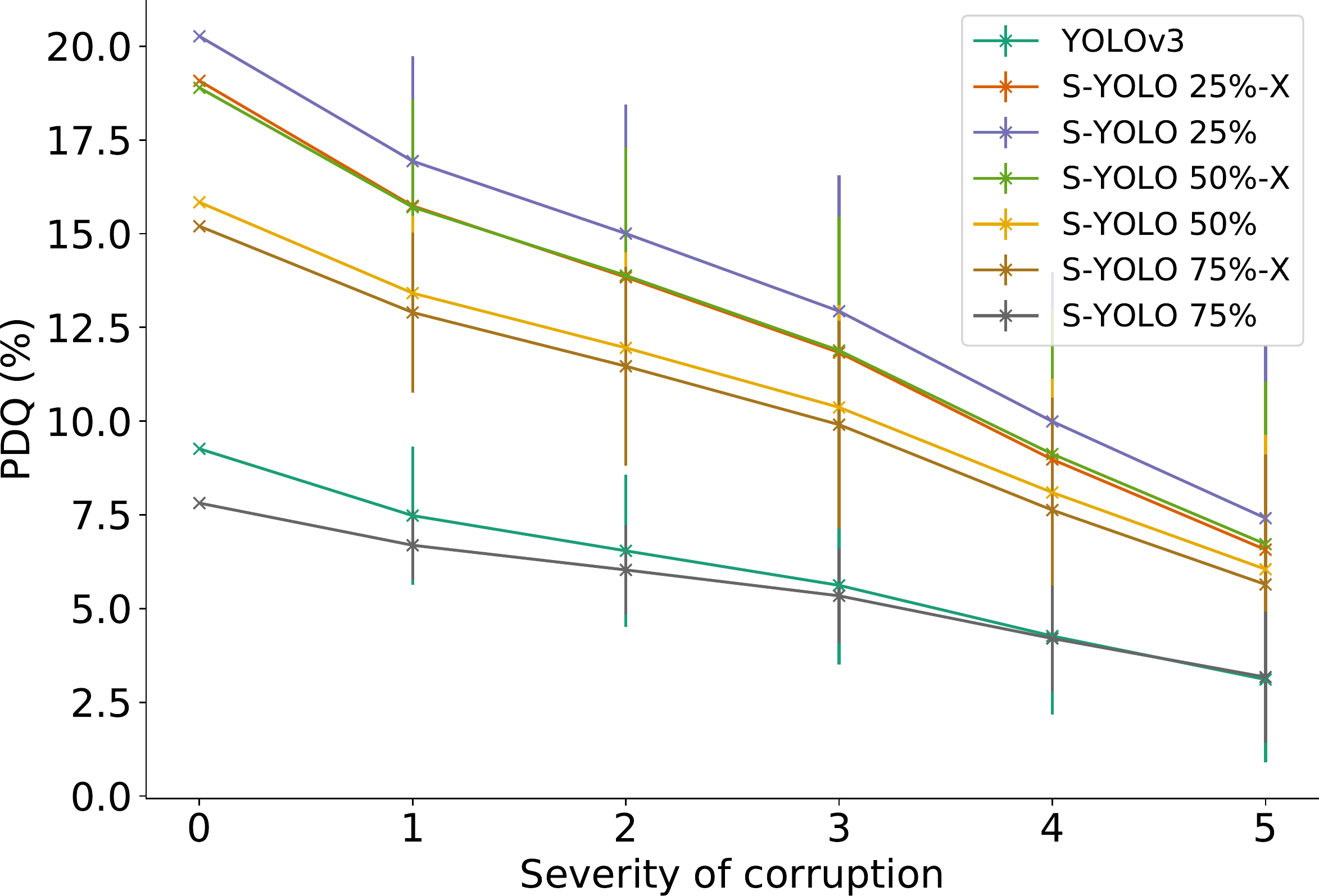}
\caption{PDQ for different models, averaged across all corruptions for each severity, with 0.5 confidence threshold. Error bars correspond to one standard deviation. S-YOLO means Stochastic-YOLO followed by dropout rate. -X means fine-tuned model.}
\label{sfig:finetune_pdq}
\end{figure}

\begin{figure}[h]
\centering
\includegraphics[width=.95\linewidth]{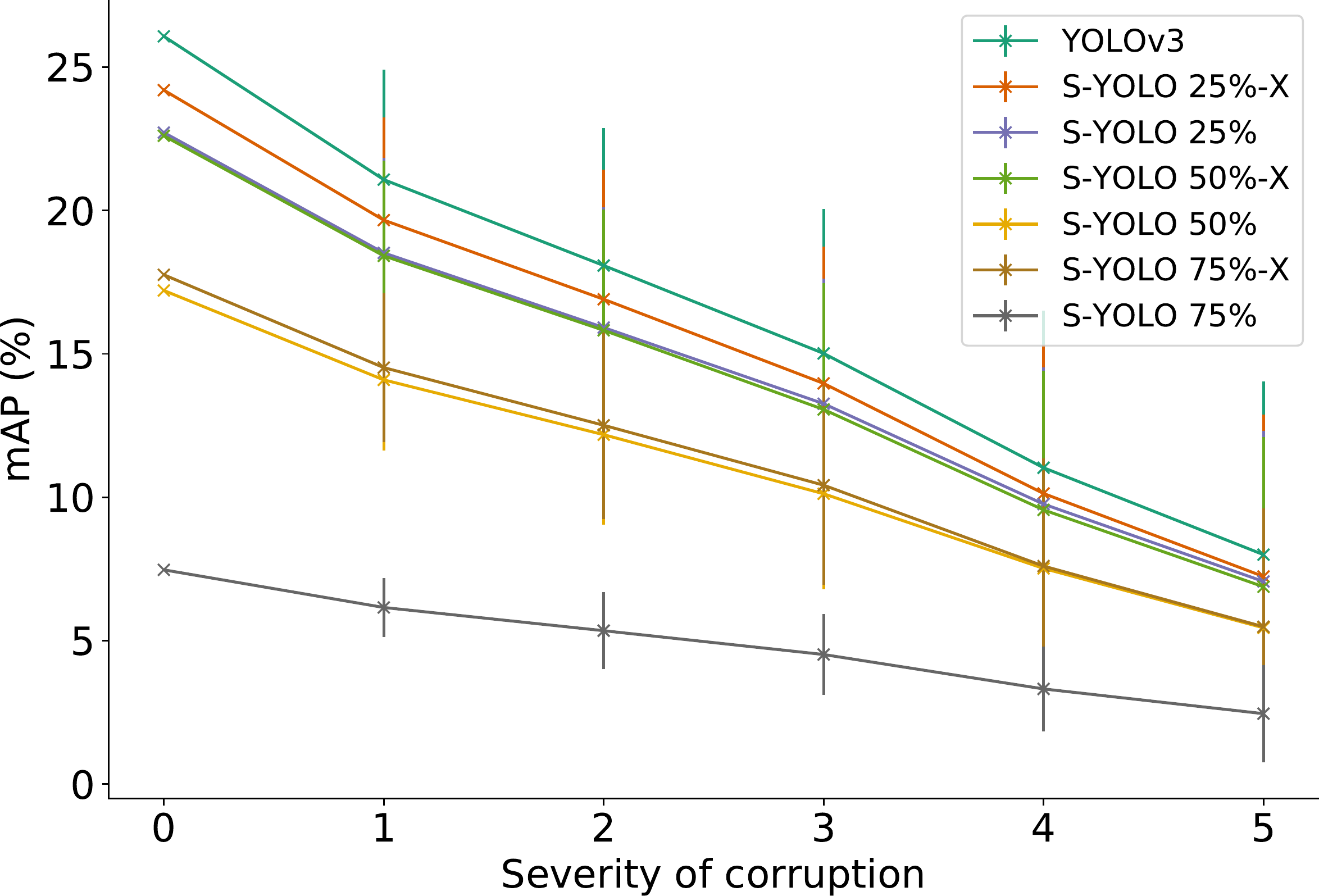}
\caption{mAP for different models, averaged across all corruptions for each severity, with 0.5 confidence threshold. Error bars correspond to one standard deviation. S-YOLO means Stochastic-YOLO followed by dropout rate. -X means fine-tuned model.}
\label{sfig:finetune_map}
\end{figure}


\begin{figure}[h]
\centering
\includegraphics[width=.95\linewidth]{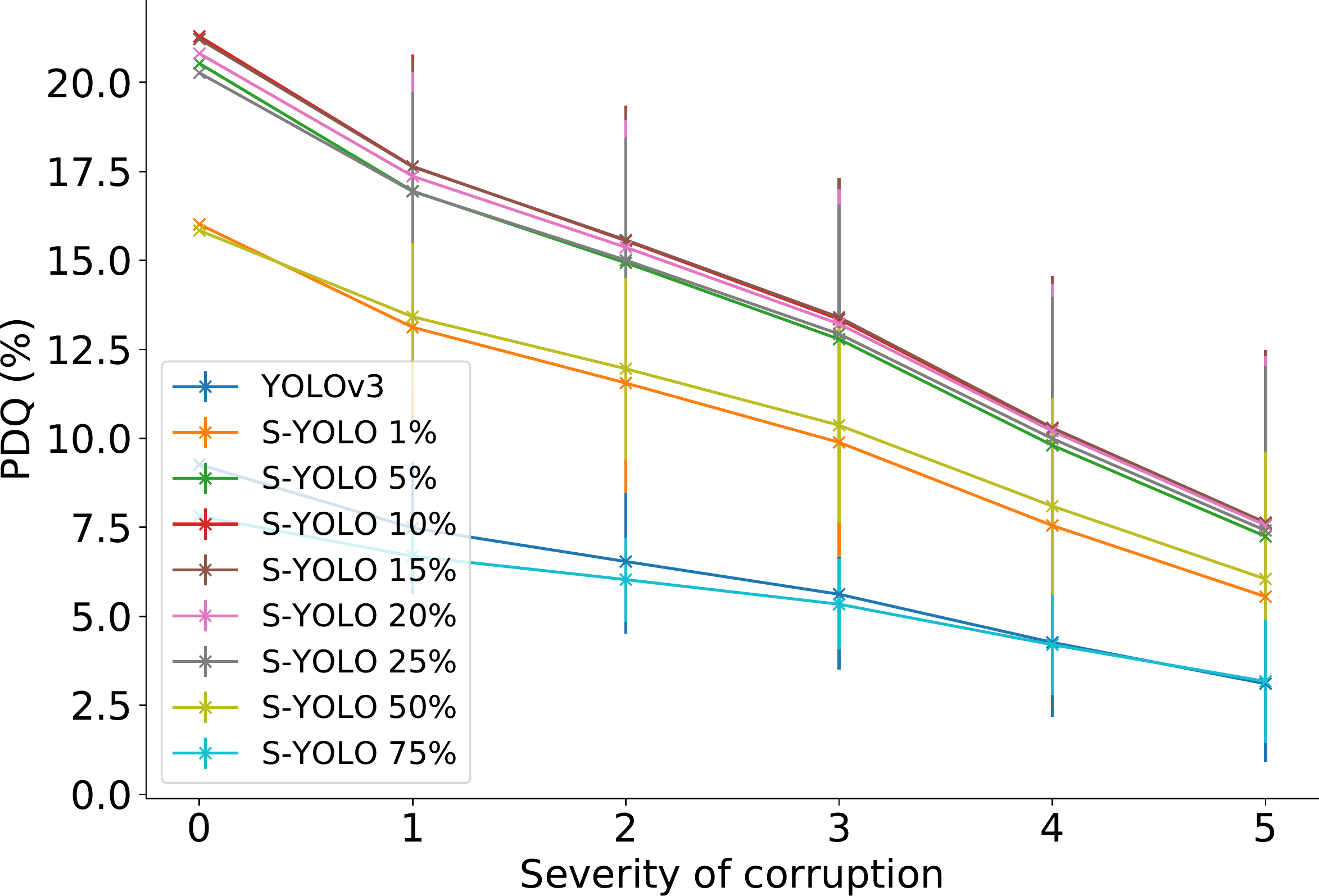}
\caption{PDQ across different dropout rates, averaged across all corruptions for each severity, with 0.5 confidence threshold. Error bars correspond to one standard deviation. S-YOLO means Stochastic-YOLO.}
\label{sfig:droprates_pdq}
\end{figure}

\begin{figure}[h]
\centering
\includegraphics[width=.95\linewidth]{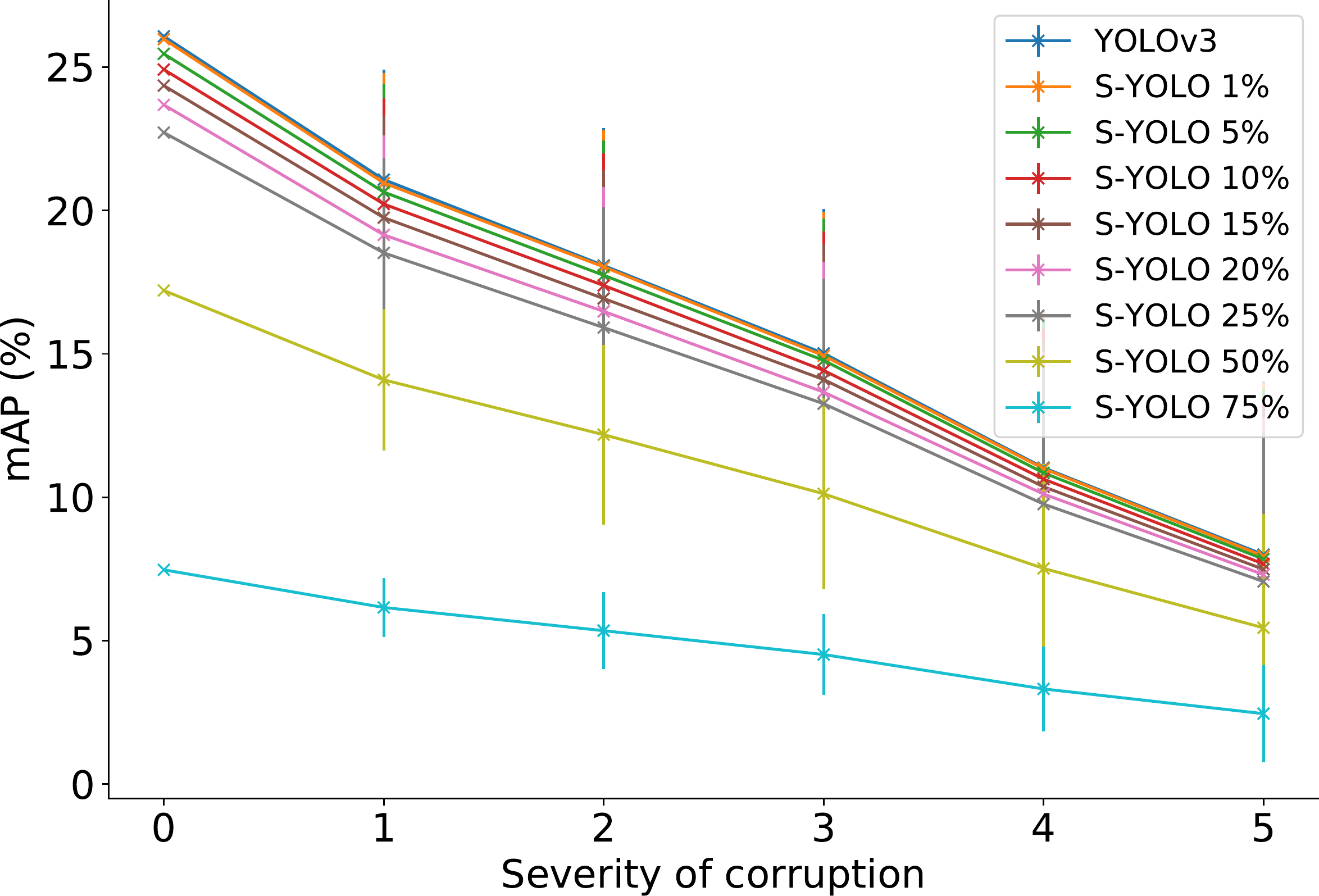}
\caption{mAP across different dropout rates, averaged across all corruptions for each severity, with 0.5 confidence threshold. Error bars correspond to one standard deviation. S-YOLO indicates Stochastic-YOLO.}
\label{sfig:droprates_map}
\end{figure}

\begin{table*}[h!]
\centering
\caption{Overall results across all evaluated cases, where Lbl and Sp mean average label and spatial uncertainty quality, respectively. In parenthesis confidence threshold. S-YOLO means Stochastic-YOLO in which the corresponding number is the dropout percentage applied, and -X means fine-tuned model}
\label{stab:all_results}
\begin{tabularx}{\textwidth}{l|XX|XX|XX|XX}
 & \textbf{mAP (\%)} & \textbf{$\textrm{rPC}_{mAP}$ (\%)} & \textbf{PDQ (\%)} & \textbf{$\textrm{rPC}_{PDQ}$ (\%)} & \textbf{Lbl (\%)} & \textbf{$\textrm{rPC}_{Lbl}$ (\%)} & \textbf{Sp (\%)} & \textbf{$\textrm{rPC}_{Sp}$ (\%)}\\
YOLOv3 (0.1) & 34.43 & 20.64 & 7.19 & 4.88 & 55.27 & 48.38 & 14.53 & 12.18 \\
YOLOv3 (0.5) & 26.07 & 14.64 & 9.26 & 5.4 & \textbf{72.88} & \textbf{69.7} & 18.42 & 16.11 \\ \hline
Ensemble-5 (0.1) & \textbf{37.14} & \textbf{22.28} & 18.36 & \textbf{13.38} & 50.66 & 43.25 & 39.24 & 37.66 \\
Ensemble-5 (0.5) & 26.0 & 14.07 & 19.53 & 11.5 & 72.21 & \textbf{69.63} & 49.21 & 48.59 \\ \hline
S-YOLO-01 (0.5) & 25.97 & 14.58 & 16.01 & 9.53 & 72.46 & \textbf{69.26} & 37.35 & 34.47 \\ \hline
S-YOLO-05 (0.5) & 25.46 & 14.37 & 20.53 & 12.34 & 71.9 & 68.82 & 48.47 & 45.62 \\ \hline
S-YOLO-10 (0.5) & 24.91 & 14.07 & \textbf{21.3} & 12.88 & 71.4 & 68.5 & 49.35 & 47.3 \\ \hline
S-YOLO-15 (0.5) & 24.35 & 13.73 & \textbf{21.22} & 12.91 & 71.01 & 68.22 & 48.73 & 46.98 \\ \hline
S-YOLO-20 (0.5) & 23.68 & 13.34 & 20.81 & 12.74 & 70.61 & 67.97 & 47.33 & 46.07 \\ \hline
S-YOLO-25 (0.1) & 31.67 & 19.31 & 17.73 & \textbf{13.4} & 48.18 & 41.67 & 37.27 & 34.84 \\
S-YOLO-25 (0.5) & 22.72 & 12.91 & 20.27 & 12.46 & 70.17 & 67.7 & 45.78 & 44.95 \\
S-YOLO-25-X (0.1) & 33.16 & 19.66 & 17.29 & 11.95 & 51.01 & 44.62 & 38.59 & 34.58 \\
S-YOLO-25-X (0.5) & 24.2 & 13.58 & 19.08 & 11.39 & 70.99 & 68.09 & 48.76 & 45.58 \\  \hline
S-YOLO-50 (0.1) & 27.37 & 16.8 & 12.04 & 10.19 & 41.47 & 35.55 & 29.39 & 27.71 \\
S-YOLO-50 (0.5) & 17.21 & 9.87 & 15.84 & 9.98 & 67.39 & 65.51 & 36.88 & 37.29 \\
S-YOLO-50-X (0.1) & 31.94 & 19.03 & 18.16 & 12.69 & 48.82 & 43.07 & 40.15 & 36.56 \\
S-YOLO-50-X (0.5) & 22.6 & 12.75 & 18.89 & 11.46 & 69.81 & 67.12 & 51.08 & 48.2 \\  \hline
S-YOLO-75 (0.1) & 17.36 & 11.11 & 4.44 & 4.2 & 29.88 & 24.79 & 19.71 & 17.79 \\
S-YOLO-75 (0.5) & 7.47 & 4.36 & 7.81 & 5.09 & 61.32 & 60.71 & 25.88 & 26.61 \\
S-YOLO-75-X (0.1) & 29.36 & 17.63 & 17.86 & 12.57 & 42.94 & 38.68 & 39.36 & 36.82 \\
S-YOLO-75-X (0.5) & 17.76 & 10.11 & 15.2 & 9.51 & 66.34 & 64.41 & \textbf{52.27} & \textbf{49.97} \\  \hline
\end{tabularx}
\end{table*}

\begin{figure*}[ht!]
\centering
\includegraphics[width=\linewidth]{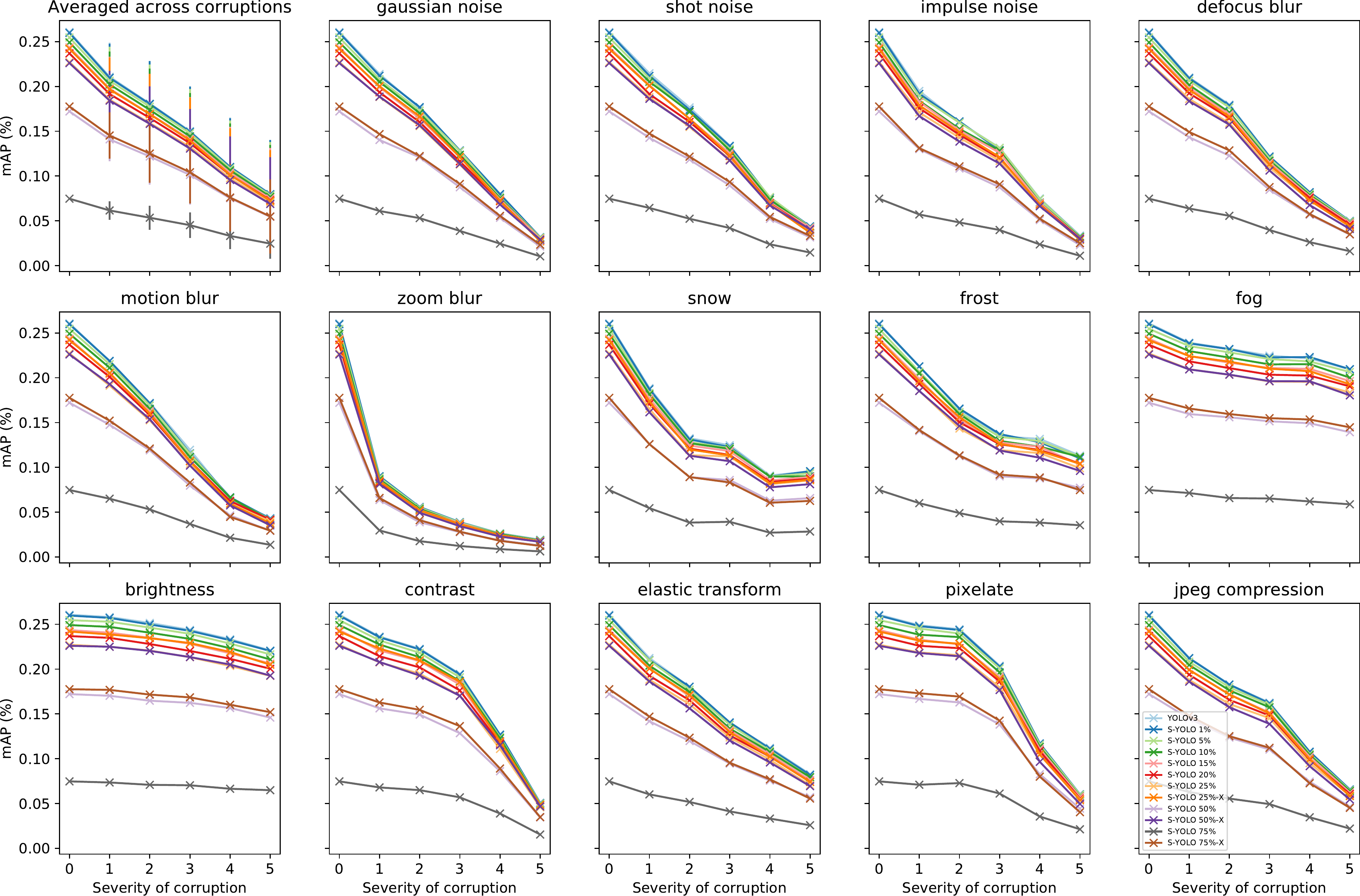}
\caption{mAP for different models on all corruption types, averaged across all corruptions for each severity in top left corner, with 0.5 confidence threshold. Error bars correspond to one standard deviation. S-YOLO means Stochastic-YOLO followed by dropout rate. -X means fine-tuned model.}
\label{sfig:suppl_corruptions_map}
\end{figure*}

\begin{figure*}[ht!]
\centering
\includegraphics[width=\linewidth]{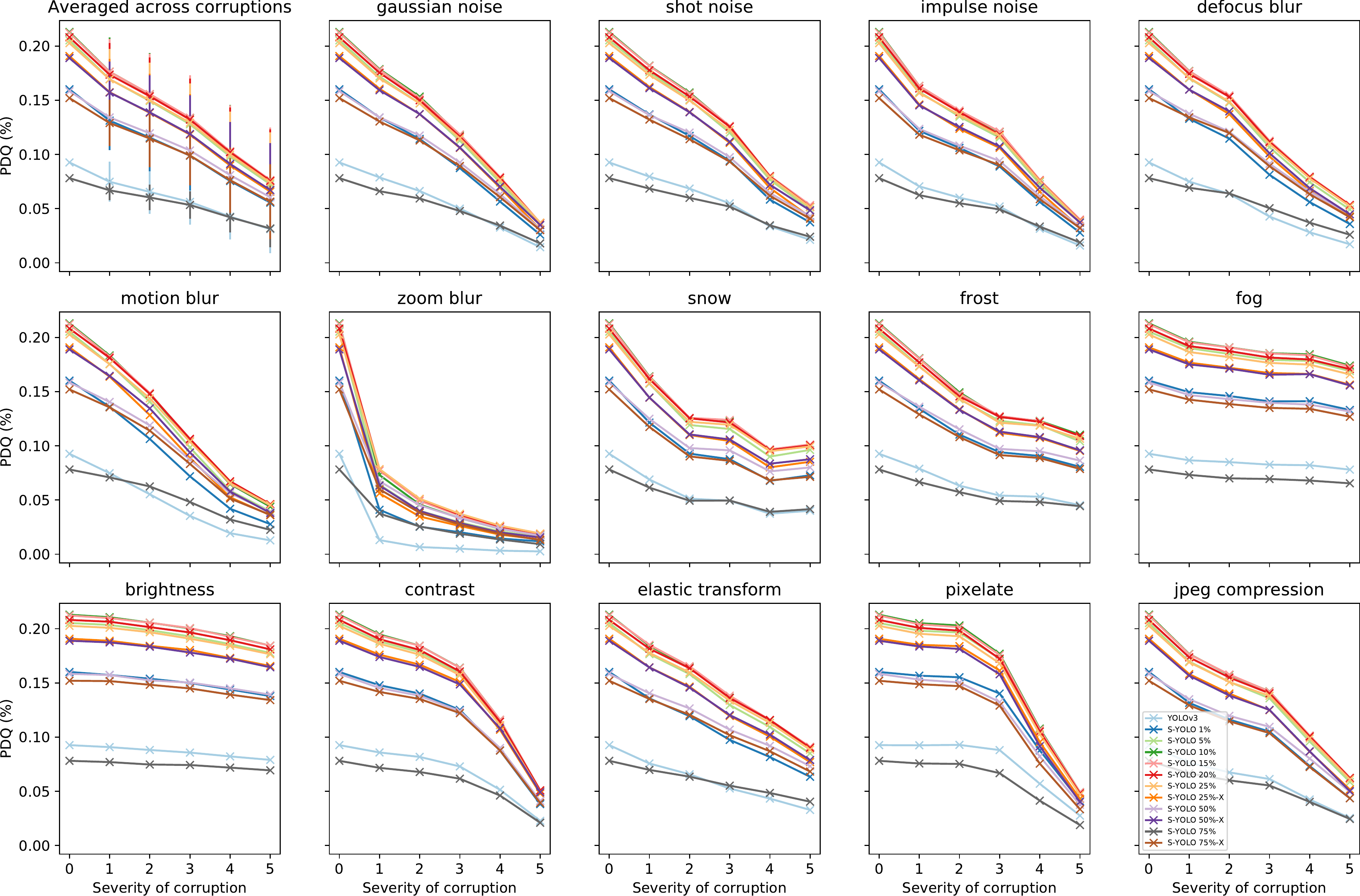}
\caption{PDQ for different models on all corruption types, averaged across all corruptions for each severity in top left corner, with 0.5 confidence threshold. Error bars correspond to one standard deviation. S-YOLO means Stochastic-YOLO followed by dropout rate. -X means fine-tuned model.}
\label{sfig:suppl_corruptions_pdq}
\end{figure*}

\begin{figure*}[ht!]
\centering
\includegraphics[width=\linewidth]{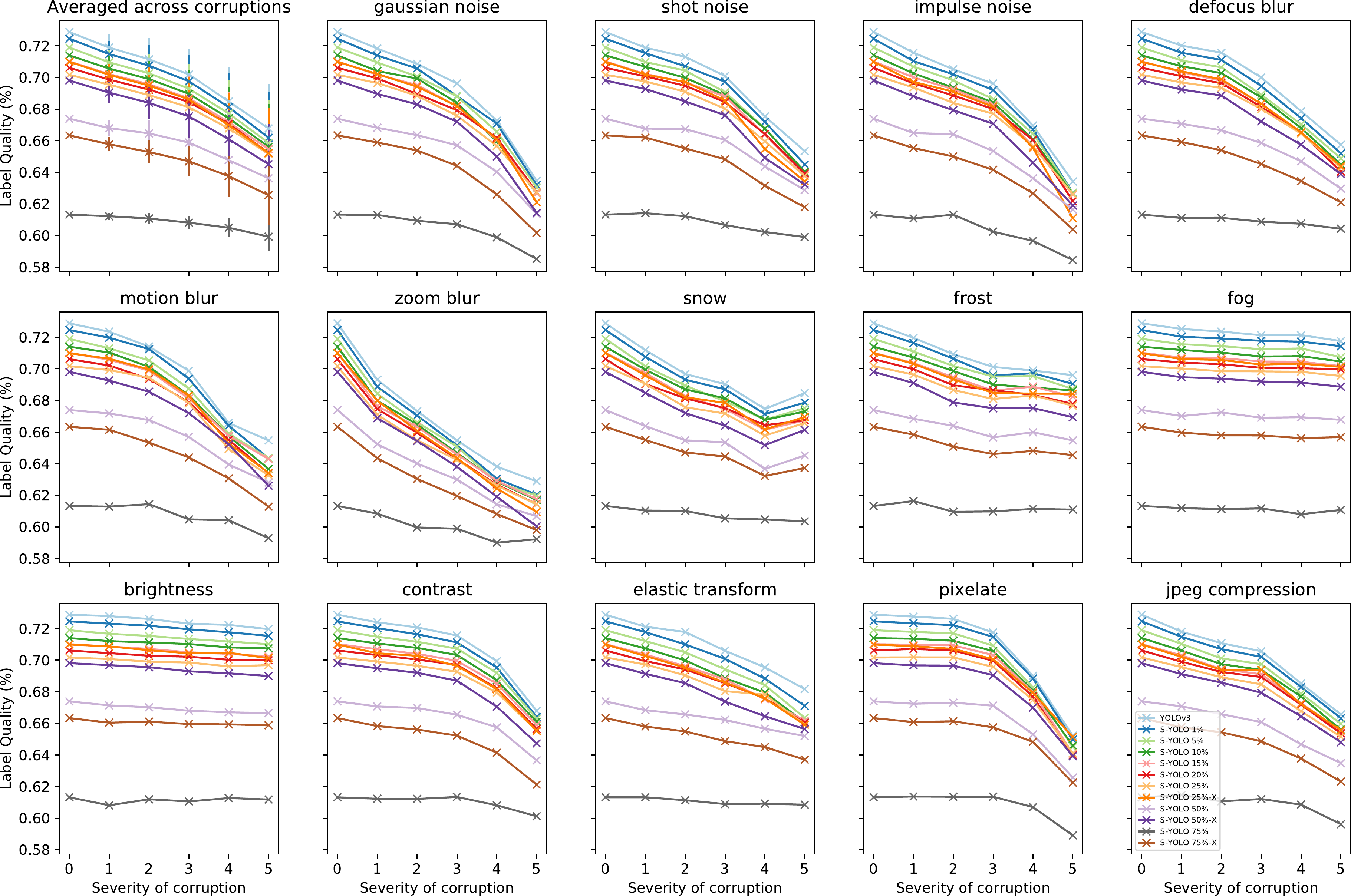}
\caption{Label Quality for different models on all corruption types, averaged across all corruptions for each severity in top left corner, with 0.5 confidence threshold. Error bars correspond to one standard deviation. S-YOLO means Stochastic-YOLO followed by dropout rate. -X means fine-tuned model.}
\label{sfig:suppl_corruptions_lbl}
\end{figure*}

\begin{figure*}[ht!]
\centering
\includegraphics[width=\linewidth]{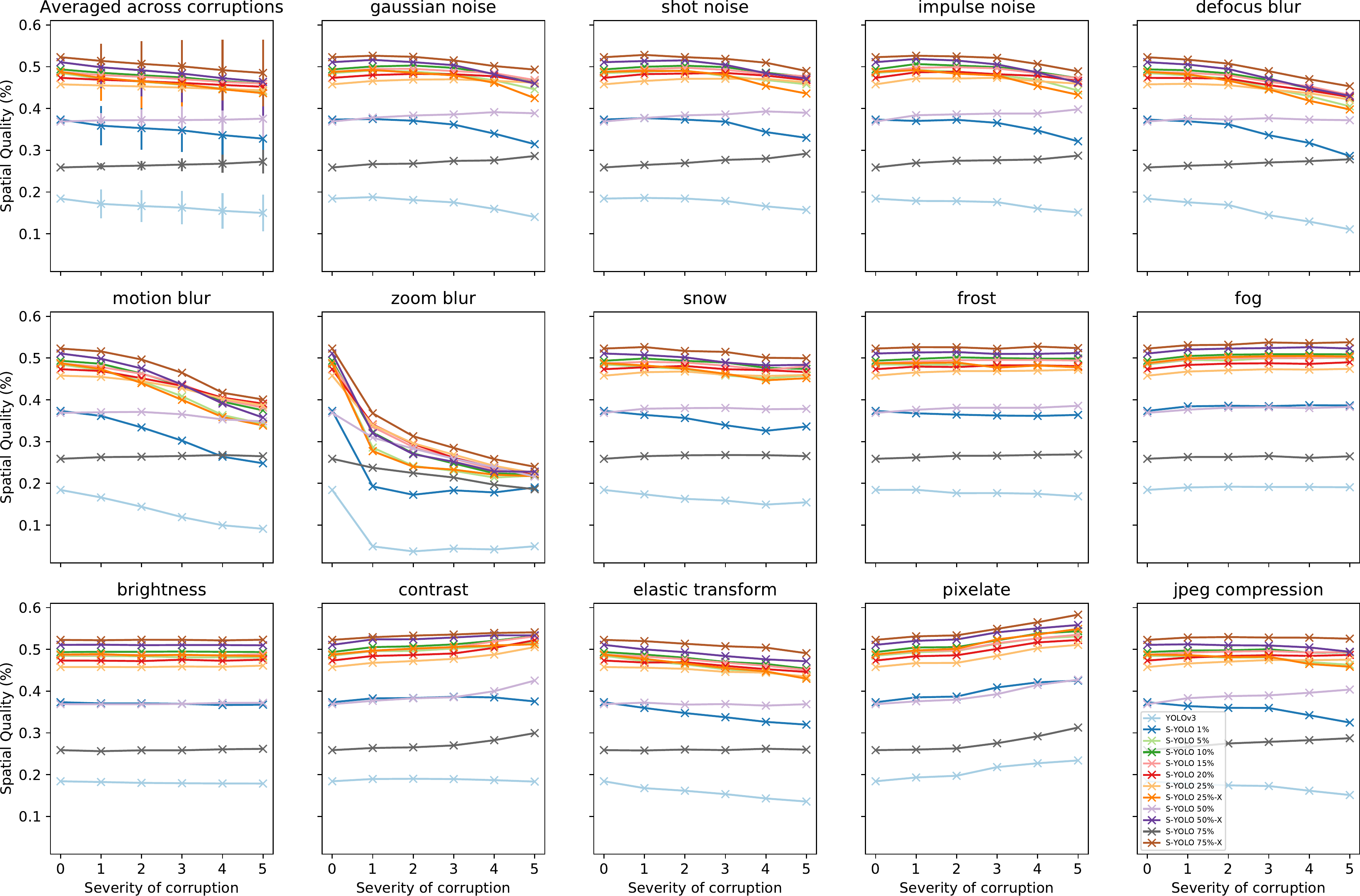}
\caption{Spatial Quality for different models on all corruption types, averaged across all corruptions for each severity in top left corner, with 0.5 confidence threshold. Error bars correspond to one standard deviation. S-YOLO means Stochastic-YOLO followed by dropout rate. -X means fine-tuned model.}
\label{sfig:suppl_corruptions_spt}
\end{figure*}
